\documentclass{article}

\PassOptionsToPackage{numbers}{natbib}

\usepackage[preprint]{neurips_2025}

\usepackage[utf8]{inputenc} 
\usepackage[T1]{fontenc}    
\usepackage{hyperref}       
\usepackage{url}            
\usepackage{booktabs}       
\usepackage{amsfonts}       
\usepackage{nicefrac}       
\usepackage{microtype}      
\usepackage{xcolor}         

\usepackage{graphicx}       
\usepackage{amsmath, amssymb}
\usepackage{caption}
\usepackage{multirow}
\usepackage{amssymb}
\usepackage{pgfplots}
\pgfplotsset{compat=1.18} 

\usepackage{array} 
\usepackage{ragged2e} 
\usepackage{graphicx} 
\newcolumntype{L}[1]{>{\RaggedRight\arraybackslash}p{#1}}

\usepackage{tabularx}
\newcommand{\intent}[1]{\texttt{#1}}
\usepackage{algorithm}
\usepackage{algpseudocode}
\usepackage{tikz}
\usetikzlibrary{shapes, arrows.meta, positioning, fit, backgrounds}
\tikzstyle{block} = [rectangle, draw=black, rounded corners=3pt,
    text centered, minimum height=2em, text width=6.3cm, font=\small, thick]
\tikzstyle{example} = [rectangle, draw=gray!60, fill=gray!10, rounded corners=2pt,
    text width=6cm, align=left, font=\footnotesize]
\tikzstyle{arrow} = [->, thick, >=stealth]
\tikzstyle{dashedbox} = [draw=gray, thick, dashed, inner sep=5pt, rounded corners=4pt]
\tikzstyle{trimbox} = [draw=blue!60, thick, dashed, inner sep=5pt, rounded corners=4pt]
\tikzstyle{dagbox} = [draw=green!60!black, thick, dashed, inner sep=5pt, rounded corners=4pt]
\tikzstyle{retrievalbox} = [draw=purple!70!black, thick, dashed, inner sep=5pt, rounded corners=4pt]

\usetikzlibrary{trees}

\title{Task Memory Engine: Spatial Memory for Robust Multi-Step LLM Agents}

\author{%
  Ye Ye\\
  New York University\\
  \texttt{yy4820@nyu.edu} \\
}
\begin{document}

\maketitle

\begin{abstract}
Large Language Models (LLMs) falter in multi-step interactions---often hallucinating, repeating actions, or misinterpreting user corrections---due to reliance on linear, unstructured context. This fragility stems from the lack of persistent memory to track evolving goals and task dependencies, undermining trust in autonomous agents. We introduce the Task Memory Engine (TME), a modular memory controller that transforms existing LLMs into robust, revision-aware agents without fine-tuning. TME implements a spatial memory framework that replaces flat context with graph-based structures to support consistent, multi-turn reasoning. Departing from linear concatenation and ReAct-style prompting, TME builds a dynamic task graph---either a tree or directed acyclic graph (DAG)---to map user inputs to subtasks, align them with prior context, and enable dependency-tracked revisions. Its Task Representation and Intent Management (TRIM) component models task semantics and user intent to ensure accurate interpretation. Across four multi-turn scenarios---trip planning, cooking, meeting scheduling, and shopping cart editing---TME eliminates 100\% of hallucinations and misinterpretations in three tasks, and reduces hallucinations by 66.7\% and misinterpretations by 83.3\% across 27 user turns, outperforming ReAct. TME’s modular design supports plug-and-play deployment and domain-specific customization, adaptable to both personal assistants and enterprise automation. We release TME’s codebase, benchmarks, and components as open-source resources, enabling researchers to develop reliable LLM agents. TME’s scalable architecture addresses a critical gap in agent performance across complex, interactive settings.
\end{abstract}

\def\wholepaper{} 
\ifdefined\wholepaper
\else
\documentclass{article}
\usepackage{neurips_2025}  
\usepackage[utf8]{inputenc} 
\usepackage[T1]{fontenc}    
\usepackage{hyperref}       
\usepackage{url}            
\usepackage{booktabs}       
\usepackage{amsfonts}       
\usepackage{nicefrac}       
\usepackage{microtype}      
\usepackage{xcolor}         
\usepackage{graphicx}       
\usepackage{amsmath, amssymb}
\usepackage{caption}
\usepackage{multirow}
\usepackage{amssymb}
\begin{document}
\fi


\section{Introduction}

LLM-based assistants excel in simple interactions but struggle with complex multi-step tasks like trip planning or meeting scheduling, often hallucinating details, repeating actions, or misinterpreting revisions, leading to \emph{contextual inconsistencies and intent misalignments}. For example, changing a travel destination to Boston may trigger redundant steps or irrelevant details, undermining trust \citep{brown2020language}. These issues stem from LLMs' reliance on linear context, which poorly tracks evolving goals, dependencies, and revisions.

Existing methods like chain-of-thought \citep{wei2022chain} and ReAct \citep{yao2022react} use linear sequences, causing context overload and semantic drift. Memory-augmented approaches \citep{zhong2023memory, wang2024augmented} often require fine-tuning or fail to handle subtask dependencies, limiting scalability. Subtle phrasing variations, like asking, "Was the meeting confirmed for Thursday?" after scheduling for Wednesday, can erroneously revise correct information.

We propose the \textbf{Task Memory Engine (TME)}, a modular memory controller enabling LLMs to reason over structured task graphs without fine-tuning. TME replaces linear context with a spatial memory framework, using the Task Memory Structure (TMS), a directed acyclic graph (DAG), to track subtasks, dependencies, and revisions. The \textbf{Task Representation and Intent Management (TRIM)} module captures semantic intent to manage input integration, branching, or revisions. By retrieving relevant memory nodes, TME ensures coherent reasoning, reducing hallucinations and saving 19.4\% tokens compared to Baseline.

\paragraph{Contributions.}
\begin{itemize}
\item \textbf{TME}: A spatial memory framework using task graphs for robust multi-step LLM interactions.
\item \textbf{TRIM}: A module for precise intent modeling and dependency tracking across turns.
\item Validation across four multi-step scenarios, achieving zero hallucinations in three tasks, 100\% reduction in hallucinations and misinterpretations over 27 turns, with open-source TME codebase and benchmarks.
\end{itemize}

TME enables scalable, reliable LLM-based interactive systems.


\ifdefined\wholepaper
\else
\end{document}
\fi

\def\wholepaper{}
\ifdefined\wholepaper
\else
\documentclass{article}
\usepackage{neurips_2025}  
\usepackage[utf8]{inputenc} 
\usepackage[T1]{fontenc}    
\usepackage{hyperref}       
\usepackage{url}            
\usepackage{booktabs}       
\usepackage{amsfonts}       
\usepackage{nicefrac}       
\usepackage{microtype}      
\usepackage{xcolor}         
\usepackage{graphicx}       
\usepackage{amsmath, amssymb}
\usepackage{caption}
\usepackage{multirow}
\usepackage{amssymb}
\begin{document}
\fi

\section{Related Work}

Prompting methods like chain-of-thought~\cite{wei2022chain}, ReAct~\cite{yao2022react}, and Tree of Thoughts~\cite{yao2023tree} guide LLMs through reasoning but suffer from context overflow or lack persistent state tracking, limiting revision handling. Memory-augmented approaches, such as MemoryBank~\cite{zhong2023memory} or retrieval-augmented generation~\cite{lewis2020rag}, store past interactions but often require fine-tuning or miss structured subtask dependencies. Graph-based and dialogue systems~\cite{heck2020trippy} handle static or simple tasks, not dynamic, revision-intensive interactions. These limitations in handling intent misalignments and revisions highlight the need for scalable memory frameworks. Unlike these, \textbf{TME} uses a spatial memory framework with dynamic task graphs to track subtasks, dependencies, and revisions without fine-tuning, ensuring scalability and consistency.

\ifdefined\wholepaper
\else
\end{document}
\fi

\def\wholepaper{}
\ifdefined\wholepaper
\else
\documentclass{article}
\usepackage{neurips_2025}  
\usepackage[utf8]{inputenc} 
\usepackage[T1]{fontenc}    
\usepackage{hyperref}       
\usepackage{url}            
\usepackage{booktabs}       
\usepackage{amsfonts}       
\usepackage{nicefrac}       
\usepackage{microtype}      
\usepackage{xcolor}         
\usepackage{graphicx}       
\usepackage{amsmath, amssymb}
\usepackage{caption}
\usepackage{multirow}
\usepackage{amssymb}
\begin{document}
\fi
\section{Problem Definition}

Large Language Models (LLMs) enable interactive agents for multi-step tasks like trip planning or meeting scheduling, but struggle with \emph{contextual inconsistencies and intent misalignments}, such as hallucinating details, retaining outdated plans, or misinterpreting queries \citep{ji2023survey, liu2024challenges}. These issues stem from linear context concatenation, which fails to filter irrelevant information or prioritize current task states.

Given a sequence of user inputs \(\{u_1, u_2, \dots, u_n\}\), where each \(u_i\) may define subtasks, revise prior inputs, or query states, the model must generate accurate responses \(\{r_1, r_2, \dots, r_n\}\) reflecting current task states and dependencies with minimal token usage. Key challenges include:
\begin{itemize}
    \item \textbf{Context Overload}: Linear context growth causes semantic drift.
    \item \textbf{Lack of Filtering}: Outdated content leads to inconsistencies.
    \item \textbf{Dependency Tracking}: Subtasks lack explicit dependency modeling.
    \item \textbf{Revision Integration}: Revisions need precise state updates.
    \item \textbf{Intent Misalignment}: Ambiguous inputs risk misinterpretation.
\end{itemize}

Unlike LLMs, human cognition uses spatial memory to organize goals \citep{baddeley2000working}. A capable system should support structured memory, selective retrieval, persistent state tracking, revision-aware reasoning, no fine-tuning, and token efficiency. We propose the \textbf{Task Memory Engine (TME)}, a spatial memory framework that dynamically constructs and filters task graphs for scalable, revision-aware reasoning in multi-turn interactions using off-the-shelf LLMs.

\ifdefined\wholepaper
\else
\end{document}
\fi

\def\wholepaper{}
\ifdefined\wholepaper
\else
\documentclass{article}
\usepackage{neurips_2025}  
\usepackage[utf8]{inputenc} 
\usepackage[T1]{fontenc}    
\usepackage{hyperref}       
\usepackage{url}            
\usepackage{booktabs}       
\usepackage{amsfonts}       
\usepackage{nicefrac}       
\usepackage{microtype}      
\usepackage{xcolor}         
\usepackage{graphicx}       
\usepackage{amsmath, amssymb}
\usepackage{caption}
\usepackage{multirow}
\usepackage{amssymb}
\newcommand{\intent}[1]{\texttt{#1}}
\usepackage{algorithm}
\usepackage{algpseudocode}
\usepackage{tikz}
\usetikzlibrary{shapes.geometric, arrows.meta, positioning}
\begin{document}
\fi

\section{Methodology}

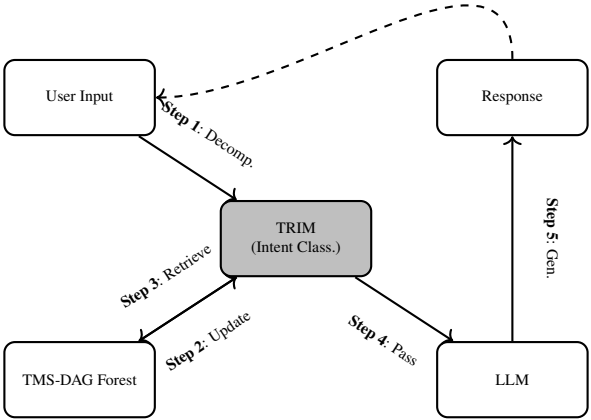
\begin{figure}[h]
    \centering
    \begin{tikzpicture}[
        scale=0.7, 
        node distance=1.2cm, 
        auto,
        every node/.style={
            rectangle, 
            draw, 
            rounded corners, 
            minimum height=1cm, 
            minimum width=2cm, 
            align=center,
            font=\tiny 
        },
        every path/.style={->, thick}
    ]
        \node[fill=lightgray] (trim) {TRIM\\(Intent Class.)};

        \node[above left=of trim] (user) {User Input};
        \node[below left=of trim] (tms) {TMS-DAG Forest};
        \node[below right=of trim] (llm) {LLM};
        \node[above right=of trim] (response) {Response};

        \draw (user) -- (trim) node[midway, above, draw=none, font=\tiny, sloped] {\textbf{Step 1}: Decomp.};
        \draw (trim) -- (tms) node[midway, below, draw=none, font=\tiny, sloped] {\textbf{Step 2}: Update};
        \draw (tms) -- (trim) node[midway, above, draw=none, font=\tiny, sloped] {\textbf{Step 3}: Retrieve};
        \draw (trim) -- (llm) node[midway, below, draw=none, font=\tiny, sloped] {\textbf{Step 4}: Pass};
        \draw (llm) -- (response) node[midway, above, draw=none, font=\tiny, sloped] {\textbf{Step 5}: Gen.};

        \draw[dashed] (response) to[out=90, in=0] (user);
    \end{tikzpicture}
    \caption{TME Architecture: TRIM orchestrates the flow through five steps: (1) decomposing inputs, (2) updating the TMS-DAG Forest, (3) retrieving subgraphs, (4) passing context to the LLM, and (5) generating responses.}
    \label{fig:tme_architecture}
\end{figure}

To address the challenges of multi-step LLM interactions---context overload, lack of information filtering, dependency tracking, revision integration, intent misalignment, and token efficiency---we propose the \textbf{Task Memory Engine (TME)}, a modular memory framework that enhances robust, revision-aware reasoning without fine-tuning. TME replaces linear context with a Directed Acyclic Graph (DAG)-based \textbf{Task Memory Structure (TMS)}, dynamically tracking subtasks, dependencies, and revisions. Inspired by human spatial memory \citep{baddeley2000working}, TME organizes task context hierarchically, enabling shared subtask reuse and global state updates. Its core component, the \textbf{Task Representation and Intent Management (TRIM)} module, decomposes complex user inputs into subtasks, classifies semantic intents, and maps them to DAG operations, ensuring \intent{contextual consistency and intent alignment}. Our implementation, validated in cooking and planning tasks, is open-source at \url{https://github.com/biubiutomato/TME-Agent}.
Figure~\ref{fig:tme_architecture} illustrates TME's architecture, with TRIM orchestrating the flow through five numbered steps.

\subsection{Task Memory Structure (TMS): From Tree to DAG}
The Task Memory Engine (TME) uses the Task Memory Structure (TMS), evolving from a tree to a Directed Acyclic Graph (DAG), defined as \( G = (V, E) \), to manage multi-step LLM interactions. Initially a tree, TMS organizes subtasks hierarchically (e.g., ``plan trip'' into ``set destination'' $\rightarrow$ ``book flight'' in Section~\ref{sec:case1}), with nodes storing \intent{slot}, \intent{value}, \intent{parent}, and \intent{history} for state tracking. As a DAG, TMS-DAG supports shared subtasks (e.g., ``wash celery'' for soup and dumplings in Section~\ref{sec:case2}), with edges \( E \) encoding dependencies (e.g., ``prepare celery'' after ``wash celery''). Operations like \intent{add} or \intent{replace} update nodes, propagating changes globally. TME manages a TMS-DAG forest for multiple tasks, enabling cross-task dependencies.

\subsection{Task Representation and Intent Management (TRIM)}

TRIM processes user input \( u_i \), which may encode multiple intents (e.g., ``make soup with celery and dumplings with tomatoes''), by decomposing it into subtasks and integrating them into the TMS-DAG forest. It classifies intents as \texttt{new} (new subtask), \texttt{update} (revise node), or \texttt{check} (query state), and infers dependencies (e.g., \texttt{depends\_on}, \texttt{replaces}). This requires nuanced reasoning to handle contextual dependencies, such as disambiguating ``no celery, use mushrooms'' as a \texttt{replace} operation.

Initially, we explored an embedding-based classifier using cosine similarity with rule-based logic, but it struggled with complex dependencies and intent disambiguation (e.g., replacing ``prepare celery'' with ``prepare mushrooms''). We adopted an LLM-based approach with few-shot prompting to output structured JSON-like representations (e.g., \{\texttt{``slot'': ``ingredient'', ``value'': ``mushrooms'', ``intent'': ``update'', ``replaces'': ``celery''}\}). In the cooking scenario, TRIM triggers a \texttt{replace} operation, updating ``prepare celery'' to ``prepare mushrooms'' across soup and dumpling tasks, ensuring consistency.

TRIM enhances efficiency by retrieving a compact subgraph \( G_i' \subseteq G \) from the TMS-DAG, minimizing context size and contributing to TME-DAG's 100\% reduction in hallucinations and confusions compared to ReAct (Table~\ref{tab:token-efficiency}). It tracks node value history for potential \texttt{roll\_back} operations (e.g., reverting to ``celery''), with full implementation planned for future work. The LLM-based prototype may be brittle in edge cases (e.g., ambiguous multi-intent inputs), a limitation we aim to address with hybrid approaches, combining rule-based filters, fine-tuned Transformers, and Graph Neural Networks for robust dependency inference.

TRIM's modular design supports extensions like \texttt{inactivate} operations to mark obsolete tasks, resolving multi-turn conflicts. This positions TRIM as a scalable component for revision-aware reasoning in TME.

\subsection{TME Workflow}
TME operates in five steps:
\ifdefined\wholepaper
\else
\documentclass{article}
\usepackage{tikz}
\usepackage{amsmath}
\usetikzlibrary{shapes, arrows.meta, positioning, fit, backgrounds}

\tikzstyle{block} = [rectangle, draw=black, rounded corners=3pt,
    text centered, minimum height=2em, text width=5.8cm, font=\footnotesize, thick]
\tikzstyle{example} = [rectangle, draw=gray!60, fill=gray!10, rounded corners=2pt,
    text width=6.3cm, align=left, font=\scriptsize]

\tikzstyle{arrow} = [->, thick, >=stealth]
\tikzstyle{dashedbox} = [draw=gray, thick, dashed, inner sep=5pt, rounded corners=4pt]
\tikzstyle{trimbox} = [draw=blue!60, thick, dashed, inner sep=5pt, rounded corners=4pt]
\tikzstyle{dagbox} = [draw=green!60!black, thick, dashed, inner sep=5pt, rounded corners=4pt]
\tikzstyle{retrievalbox} = [draw=purple!70!black, thick, dashed, inner sep=5pt, rounded corners=4pt]

\begin{document}
\fi

\begin{figure}[ht]
\centering
\scalebox{0.85}{
\hspace*{-1.8cm} 
\begin{tikzpicture}[node distance=0.5cm and 0.6cm]

\node (input) [block] {\textbf{User Input:} $u_i$, TMS-DAG $G$};
\node (step1) [block, below=of input] {1. Input Decomposition:\\ TRIM splits $u_i$ into $S$};
\node (step2) [block, below=of step1] {2. Intent Classification:\\ $z_j = f_{\mathrm{TRIM}}(s_j, G)$};
\node (step3) [block, below=of step2] {3. TMS-DAG Update:\\ $G_{i+1} = \mathrm{Update}(G_i, S, \{z_j\})$};
\node (step4) [block, below=of step3] {4. Context Retrieval:\\ $G'_i = \arg\min \sum \mathrm{cost}(n)$};
\node (step5) [block, below=of step4] {5. Response Generation:\\ $r_i = \mathrm{LLM}(u_i, G'_i)$};
\node (output) [block, below=of step5] {\textbf{Output:} $G_{i+1}, r_i$};


\node (ex0) [example, right=of input] {“no celery, use mushrooms, search for a cake recipe”, TMS with “prepare celery” node};
\node (ex1) [example, right=of step1] {Split into: \\ Task 1: “no celery, use mushrooms” (substitution) \\ Task 2: “search for a cake recipe” (recipe request)};
\node (ex2) [example, right=of step2] {Classified as: \\ Task 1: \texttt{update} intent \\ Task 2: \texttt{new} intent};
\node (ex3) [example, right=of step3] {Task 1: Replace “prepare celery” with “prepare mushrooms” \\ Task 2: Add intent to retrieve a cake recipe};
\node (ex4) [example, right=of step4] {Task 1: Substitution updated in TMS-DAG \\ Task 2: Retrieve subgraph for a cake recipe};
\node (ex5) [example, right=of step5] {"Task 1: Celery replaced with mushrooms in TMS-DAG, affecting preparation of relevant dishes like soups or dumplings. \\ Task 2: Here’s the cake recipe.”};
\node (ex6) [example, right=of output] {Updated TMS-DAG + Generated response};

\draw [arrow] (input) -- (step1);
\draw [arrow] (step1) -- (step2);
\draw [arrow] (step2) -- (step3);
\draw [arrow] (step3) -- (step4);
\draw [arrow] (step4) -- (step5);
\draw [arrow] (step5) -- (output);

\node[fit=(step1)(step2), trimbox, label=left:{\textbf{TRIM}}] {};
\node[fit=(step3), dagbox, label=left:{\textbf{TMS-DAG Update}}] {};
\node[fit=(step4)(step5), retrievalbox, label=left:{\textbf{Retrieval + Response}}] {};

\end{tikzpicture}
}
\caption{TME execution pipeline (left) with example trace from the cooking scenario (right). 
\textbf{TRIM} handles input decomposition and intent classification (Steps 1–2); 
\textbf{TMS-DAG Update} occurs in Step 3; 
\textbf{Retrieval + Response} are Steps 4–5.}
\label{fig:tme_example_pipeline}
\end{figure}
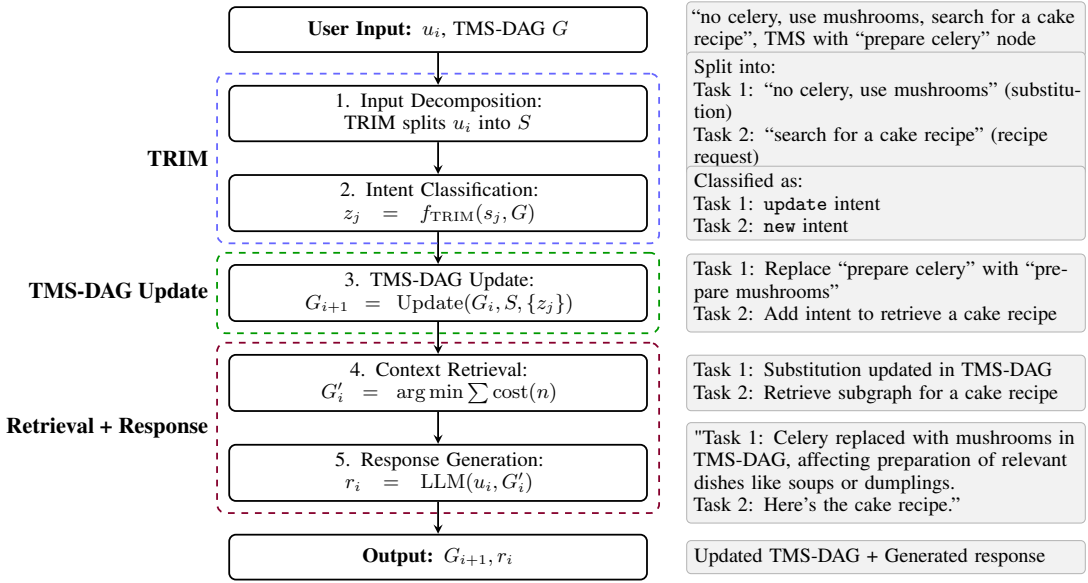

\ifdefined\wholepaper
\else
\end{document}
\fi

\begin{enumerate}
    \item \textbf{Input Decomposition}: TRIM splits \( u_i \) into subtasks (e.g., ``wash celery'', ``chop celery'') using an LLM prompt, as in Algorithm~\ref{alg:workflow}.
    \item \textbf{Intent Classification}: TRIM classifies each subtask’s intent (\intent{new}, \intent{update}, \intent{check}) and infers relationships (e.g., \intent{replaces} for ``celery'' to ``mushrooms'').
    \item \textbf{TMS-DAG Update}: TME updates the TMS-DAG forest, adding nodes for new subtasks, replacing nodes for updates, deleting nodes for removed subtasks, rolling back nodes to prior states for revisions, or linking dependencies across task trees.
    \item \textbf{Context Retrieval}: TRIM retrieves a compact subgraph \( G_i' \subseteq G \) from the relevant TMS-DAG within the forest, minimizing token usage.
    \item \textbf{Response Generation}: The LLM generates response \( r_i \) using \( G_i' \), ensuring consistency.
\end{enumerate}
For example, in the cooking case study, ``no celery, use mushrooms'' triggers TRIM to replace the shared ``prepare celery'' node, update dependencies (e.g., soup and dumpling preparation), retrieve relevant nodes, and generate a revised recipe, avoiding outdated context.

\begin{algorithm}
\caption{TME Workflow for Subtask Decomposition and TMS-DAG Update}
\label{alg:workflow}
\begin{algorithmic}[1]
\State \textbf{Input}: User input \( u_i \), DAG \( G = (V, E) \)
\State \textbf{Output}: Updated TMS-DAG \( G_{i+1} \), response \( r_i \)
\State \( S \gets \text{Decompose}(u_i) \) \Comment{LLM-based subtask splitting}
\For{\( s_j \in S \)}
    \State \( \mathbf{z}_j \gets \text{ClassifyIntent}(s_j, G) \) \Comment{Intent: new, update, check}
    \If{\( \mathbf{z}_j = \intent{new} \)}
        \State \( n_j \gets \text{TaskNode}(s_j.\text{slot}, s_j.\text{value}, \text{parent}, \text{deps}) \)
        \State \( V \gets V \cup \{n_j\} \), \( E \gets E \cup \text{InferEdges}(n_j, G) \)
    \ElsIf{\( \mathbf{z}_j = \intent{update} \)}
        \State \( n_k \gets \text{FindNode}(s_j.\text{slot}, G) \)
        \State \( n_k.\text{value} \gets s_j.\text{value} \), \( n_k.\text{history} \gets n_k.\text{history} \cup \{n_k.\text{value}\} \)
        \State \( E \gets \text{PropagateDeps}(n_k, G) \)
    \Else
        \State \( G' \gets \text{RetrieveSubgraph}(s_j, G) \)
        \State \( r_j \gets \text{LLM}(s_j, G') \)
    \EndIf
\EndFor
\State \( G_{i+1} \gets (V, E) \), \( r_i \gets \text{LLM}(u_i, G') \)
\State \textbf{return} \( G_{i+1} \), \( r_i \)
\end{algorithmic}
\end{algorithm}

\subsection{Addressing Challenges}
TME addresses the challenges outlined in Section 2:
\begin{itemize}
    \item \textbf{Context Management}: TMS-DAG’s subgraph retrieval minimizes context size and filters outdated content, unlike linear prompts \citep{yao2022react}.
    \item \textbf{Dependency and Revision Handling}: TMS-DAG edges track dependencies, and TRIM’s \intent{replace} operation ensures global updates.
    \item \textbf{Intent and Efficiency}: TRIM’s intent classification disambiguates inputs, achieving 19.4\% token savings (see Evaluation).
\end{itemize}

\subsection{Formal Description}
For input \( u_i \), TRIM decomposes it into subtasks \( S = \{s_1, s_2, \dots\} \), computing intent vectors \( \mathbf{z}_j = f_{\text{TRIM}}(s_j, G) \) for each \( s_j \), where \( f_{\text{TRIM}} \) is an embedding-based classifier mapping to \{\intent{new}, \intent{update}, \intent{check}\}. The DAG update is:
\[
G_{i+1} = \text{Update}(G_i, S, \{\mathbf{z}_j\}),
\]
where \(\text{Update}\) adds nodes (\( V \gets V \cup \{n_j\} \)), updates values (\( n_k.\text{value} \gets s_j.\text{value} \)), or propagates dependencies. Context retrieval selects a subgraph:
\[
G_i' = \arg\min_{G' \subseteq G} \sum_{n \in G'} \text{cost}(n),
\]
passed to the LLM as \( r_i = \text{LLM}(u_i, G_i') \). This ensures compact, relevant prompts, minimizing token usage.

\subsection{Implementation Notes}
TME is implemented as a lightweight layer compatible with off-the-shelf LLMs (e.g., GPT-4o), using memory-efficient DAG storage (e.g., adjacency lists). TRIM’s classifier leverages an LLM-based approach for contextual intent classification and dependency inference, ensuring scalability and robust reasoning across multi-turn interactions. A large language model aided in designing experiments and debugging code, with outputs reviewed by the research team to ensure alignment with study goals, following AI tool usage disclosure guidelines.


\ifdefined\wholepaper
\else
\end{document}
\fi

\def\wholepaper{}
\section{Case Studies}
\label{sec:casestudies}
We evaluate the \textbf{Task Memory Engine (TME)} across four scenarios---trip planning, cooking, meeting scheduling, and cart editing---comparing its DAG-based memory (TME-DAG) against ReAct-style flat memory \citep{yao2022react}, focusing on ReAct’s hallucinations and confusions that TME avoids. Each case study tests TME’s ability to handle multi-step tasks with revisions, ensuring \emph{contextual consistency and intent alignment}.

\subsection{Case I: Trip Planning with Destination Revisions}
\label{sec:case1}

\paragraph{Script Overview.} The user plans a trip across 10 rounds: setting the destination to Seattle, start from Chicago, departure on June 10th, revising the destination to San Francisco, reverting to Seattle, querying the start (``Wasn’t I departing from Boston?''), and requesting flights and a summary.

\vspace{-0.5em}
\begin{figure}[ht]
\centering
\hspace*{-2cm}  
\includegraphics[width=1.3\linewidth]{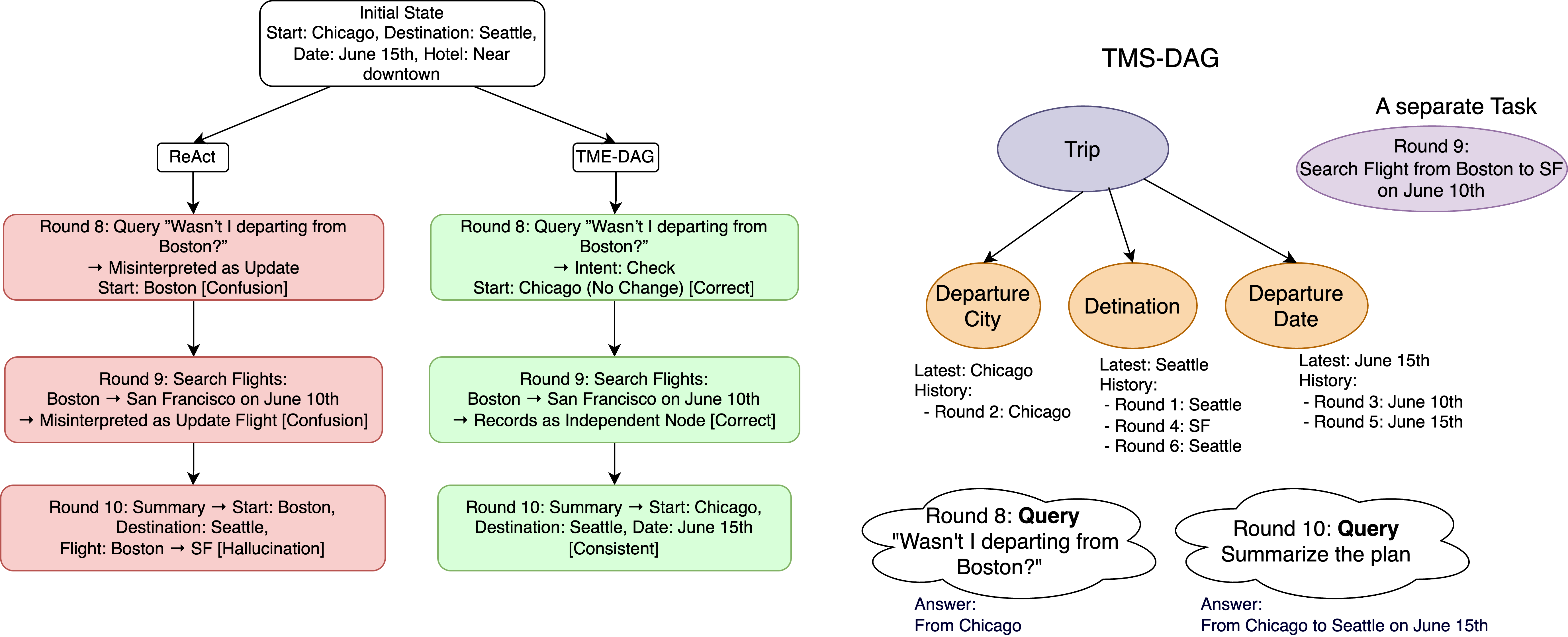}  
\caption{
ReAct vs. TME-DAG: Hallucination and Confusion in Trip Planning Scenario (Rounds 8–10). This diagram illustrates how ReAct misinterprets a user query in Round 8 and a flight search in Round 9 as updates, resulting in hallucination in Round 10. In contrast, TME-DAG correctly treats the query as a check, logs the new flight as an independent node, and produces a consistent summary via memory tracking and slot-based dependency reasoning.
}
\label{fig:trip-case-study}
\end{figure}
\vspace{-1em}

\paragraph{Failure Modes in ReAct.} ReAct hallucinates in Round 10: after updating the flight to ``Boston to San Francisco on June 10th'' (Round 9), it summarizes with ``Destination: Seattle, Flight: Boston to San Francisco'' (Seattle \(\neq\) San Francisco), as its linear context fails to reconcile destinations. In Round 8, ReAct misinterprets the query as an update, changing the start to Boston from Chicago.

\begin{itemize}
    \item \textbf{Hallucination:} Final summary conflicts with user inputs: \textit{Destination = Seattle} vs. \textit{Flight = Boston \(\rightarrow\) SF}.
    \item \textbf{Confusion:} Misinterprets a clarification question as an update (Round 8).
\end{itemize}

\paragraph{Success Modes in TME-DAG.} TME-DAG creates a task graph within the TMS-DAG forest, with nodes for destination, start, and flight. In Round 10, it prioritizes the destination node (Seattle), responding: ``Destination: Seattle, Start: Chicago, Date: June 15th.'' TRIM classifies the Round 8 query as a \emph{check} intent, preserving the correct start (Chicago) and replying: ``I couldn't find evidence that you were departing from Boston.''

\subsection{Case II: Cooking with Cross Dependencies}
\label{sec:case2}

\paragraph{Script Overview.} The user prepares soup and dumplings over 7 rounds: (1) ``To make soup, wash and chop celery''; (2) ``To make dumplings, chop tomatoes and peel \& chop shrimp''; (3) ``Also use celery in dumplings''; (4) ``Wait! There's no celery in the refrigerator at all. Let's all use mushrooms instead''; followed by queries about ingredients and celery’s inclusion.

\begin{figure}[ht]
\centering
\includegraphics[width=1.0\linewidth]{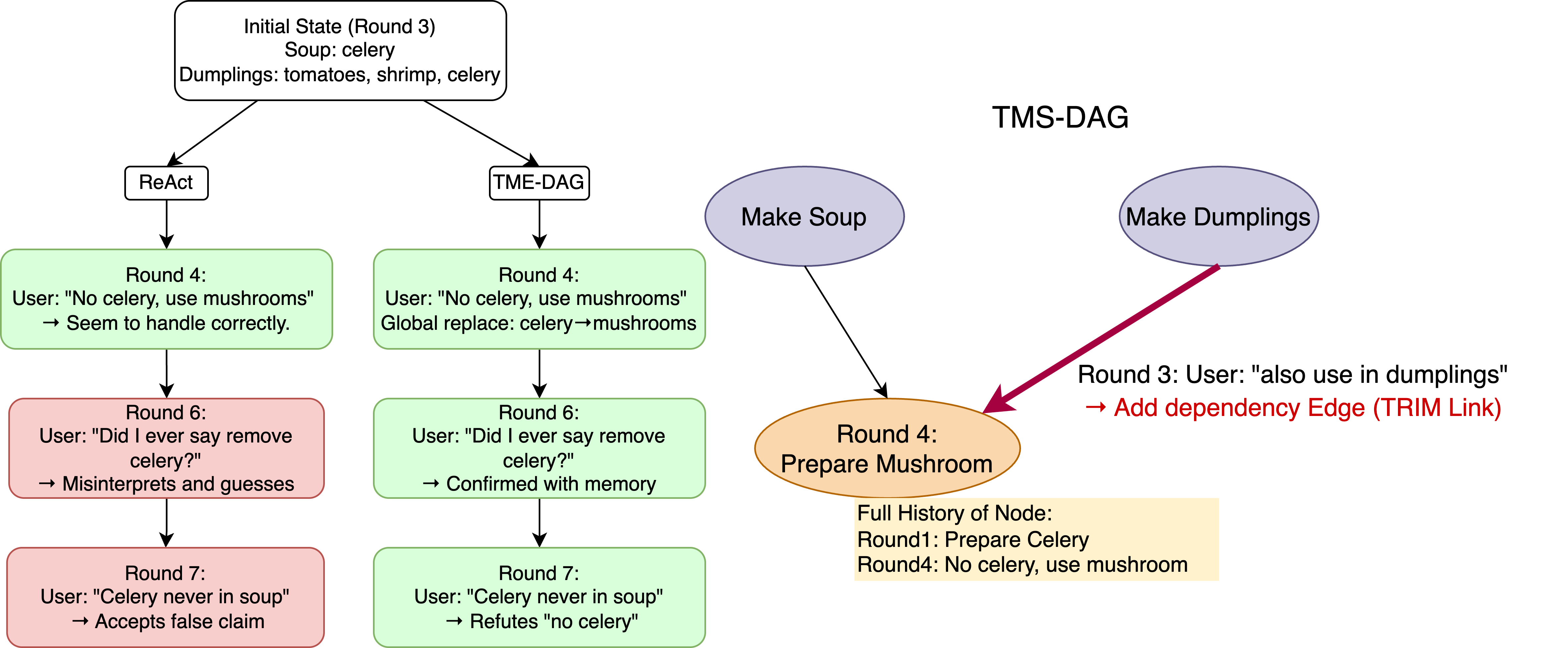}  
\caption{Comparison of ReAct and TME-DAG in a cooking scenario with ingredient substitution. Left: ReAct exhibits memory inconsistencies. Right: TME-DAG ensures cross-task consistency via graph-based updates.}
\label{fig:cook-case-study}
\end{figure}

\paragraph{Failure Modes in ReAct.} ReAct hallucinates in Round 7 (``I think celery was never part of the soup''), accepting the incorrect assumption and failing to validate celery’s initial inclusion, due to its linear history. In Round 6, it misinterprets the celery-to-mushroom substitution as a potential removal in dumplings, despite the global replacement instruction.

\begin{itemize}
    \item \textbf{Hallucinations:} Accepts erroneous assumption about soup (Round 7).
    \item \textbf{Confusions:} Misinterprets substitution as removal in dumplings (Round 6).
\end{itemize}

\paragraph{Success Modes in TME-DAG.} TME-DAG creates a shared ``prepare celery'' node across the TMS-DAG forest for soup and dumplings. The substitution (``use mushrooms instead'') triggers an \intent{replace} operation, updating to ``prepare mushrooms'' globally. In Round 6, it clarifies: ``Celery was replaced with mushrooms for all dishes.'' In Round 7, it validates: ``Celery was initially in the soup, replaced in Round 4.''

\subsection{Case III: Meeting Scheduling with Participant Conflicts}

\paragraph{Script Overview.} The user schedules a meeting over 5 rounds: adjusting time for Carol, splitting for Bob, consolidating to 3 PM, and querying the schedule.

\paragraph{Failure Modes in ReAct.} ReAct v2 hallucinates in Round 4 (Make it a single 3 PM meeting''), omitting Carol due to misinterpreting Round 2 (Carol can’t make 2 PM'') as permanent exclusion. It also misinterprets the split request in Round 3, continuing to exclude Carol.
\begin{itemize}
\item \textbf{Hallucination:} Omits Carol from 3 PM meeting (Round 4).
\item \textbf{Confusions:} (1) Permanent exclusion of Carol (Round 2); (2) persists exclusion in split meeting (Round 3).
\end{itemize}

\paragraph{Success Modes in TME-DAG.} TME-DAG tracks participant nodes linked to the schedule, updating time to 4 PM, splitting correctly (Bob at 2–2:45, Alice and Carol at 4 PM), and consolidating to 3 PM with all participants, ensuring consistency.

\subsection{Case IV: Cart Editing with Item Conflicts}
\paragraph{Script Overview (Cart Editing).}

The user manages a shopping cart over five turns: (1) adding two iPhone cases (black, clear), a charger, and a MacBook stand; (2) removing the clear case and charger; (3) correcting to keep the charger and remove the black case; (4) querying cart contents; (5) resetting to MacBook stand and charger only.

\paragraph{ReAct Behavior.} ReAct accurately tracks item states using linear context, correctly processing additions, removals, corrections, and queries, finalizing with MacBook stand and charger. It incurs \textbf{zero hallucinations and confusions}, ideal for flat, sequential tasks.

\paragraph{TME-DAG Behavior.} Initially, TME-DAG struggled due to its hierarchical design for complex tasks. In Round 4, it misreported “two iPhone cases and MacBook stand,” incurring \textbf{1 hallucination} and \textbf{1 confusion} from residual dependency conflicts in the DAG, as inactive iPhone case nodes persisted. We introduced a \textbf{task-specific TRIM adaptation}, flattening the structure to treat items as independent nodes with direct state updates. Post-adaptation, TME-DAG achieved \textbf{zero hallucinations and confusions}, matching ReAct’s performance.

\paragraph{Takeaways.} ReAct suits flat tasks, while TME-DAG’s initial complexity caused errors in cart editing. The \textbf{plug-and-play TRIM adaptation} highlights TME-DAG’s flexibility: task-specific components (e.g., flattened TRIM) integrate seamlessly, aligning its memory structure with flat tasks while retaining structured reasoning for dependency-heavy scenarios, ensuring robust performance across diverse tasks.

\begin{table}[h]
\centering
\small
\caption{Performance across case studies (27 user turns).}
\label{tab:casestudy}
\begin{tabular}{lcccc}
\toprule
\textbf{System} & \textbf{Rounds} & \textbf{Hallucinations} & \textbf{Confusions} & \textbf{Consistent Tasks} \\
\midrule
base-flat   & 27 & 4 & 4 & 2/4 \\
CoT         & 27 & 4 & 1 & 3/4 \\
ReAct       & 27 & 3 & 5 & 2/4 \\
TME-DAG     & 27 & 0 & 0 & 4/4 \\

\bottomrule
\end{tabular}
\end{table}

\section{Experiments}
\label{sec:experiments}
Experiments used ChatGPT-4o as the underlying LLM, ensuring compatibility with TME’s lightweight layer for robust multi-turn reasoning. We evaluate TME-DAG across 27 rounds in four scenarios (trip planning, cooking, meeting scheduling, cart), focusing on aggregated metrics, ablation studies, and scalability, complementing qualitative insights from Section 5. TME-DAG is compared against ReAct \citep{yao2022react}, a standard baseline for multi-step reasoning, alongside base-flat and CoT for a comprehensive evaluation.

\subsection{Aggregated Performance Metrics}


Table~\ref{tab:casestudy} summarizes performance across all scenarios. TME-DAG reduces hallucinations by 100\% (0 vs. 3) and confusions by 100\% (0 vs. 5) compared to ReAct, achieving 100\% task consistency (4/4 tasks).

\subsection{Token Efficiency}

We compare TME-DAG’s token usage to a baseline with flat memory (base-flat) in a 6-turn form-filling task. TME-DAG uses TRIM to prune irrelevant nodes from the TMS-DAG, retrieving only relevant subgraphs, unlike base-flat’s full history concatenation. TME-DAG reduces total tokens by 19.4\% (725 vs. 899) over 6 rounds and 26.4\% (673 vs. 899) over the first five, as shown in Table~\ref{tab:token-efficiency}. TRIM’s filtering ensures compact, relevant prompts, enabling longer interactions within LLM context limits and enhancing scalability. Additional scenario comparisons (e.g., trip planning, cooking) are in the Appendix.

\begin{table}[h]
\centering
\begin{tabular}{lccc}
\toprule
\textbf{Task} & \textbf{System} & \textbf{Total Tokens} & \textbf{Savings} \\
\midrule
Form-Filling & base-flat & 899 & -- \\
Form-Filling & TME-DAG & \textbf{725} & 19.4\% \\
\bottomrule
\end{tabular}
\caption{Token usage comparison for the form-filling task. See appendix for detailed per-round token usage.}
\label{tab:token-efficiency}
\end{table}

\subsection{Ablation Studies}

We conducted ablation studies on TME-DAG across 27 user turns in four multi-step scenarios (trip planning, cooking, meeting scheduling, cart editing) from Section~\ref{sec:casestudies}, using ChatGPT-4o (Section~\ref{sec:experiments}). Two variants were tested: \textbf{TME-RandomTRIM}, which disables the TRIM module by replacing LLM-based intent classification with random assignments (\texttt{new}, \texttt{update}, \texttt{check}), and \textbf{TME-Flat}, which disables TMS-DAG, using ReAct-style linear context concatenation. TME-RandomTRIM causes errors, e.g., misinterpreting ``no celery, use mushrooms'' as a check, omitting mushrooms (Round 5, Section~\ref{sec:case2}), or altering trip start locations (Round 8, Section~\ref{sec:case1}). TME-Flat loses dependency tracking, failing to propagate ``celery $\to$ mushrooms'' in cooking or causing conflicting destinations in trip planning (Section~\ref{sec:case1}). Table~\ref{tab:ablation} shows TME-RandomTRIM yields 3 hallucinations, 6 confusions, and 0/4 consistent tasks; TME-Flat has 2 hallucinations, 4 confusions, and 1/4 consistent tasks; TME-DAG achieves 4/4 consistent tasks. See Appendix for examples.

\begin{table}[t]
\centering
\caption{Ablation study results across 27 user turns. }
\label{tab:ablation}
\begin{tabular}{lccc}
\toprule
System & Hallucinations & Confusions & Consistent Tasks \\
\midrule
TME-DAG & 0 & 0 & 4/4 \\
TME-RandomTRIM & 3 & 6 & 0/4 \\
TME-Flat & 2 & 4 & 1/4 \\
ReAct & 3 & 5 & 2/4 \\
\bottomrule
\end{tabular}
\end{table}

\section{Conclusion and Future Work}

The Task Memory Engine with Directed Acyclic Graphs (TME-DAG) transforms LLM reliability in multi-step, interactive tasks by replacing linear context with a dynamic, graph-based memory tree. This spatial framework ensures global task consistency and revision-aware reasoning, eliminating hallucinations and intent misalignments. Across 27 user turns in four scenarios---trip planning, cooking, meeting scheduling, and cart editing---TME-DAG achieves 100\% reductions in hallucinations and confusions, outperforming ReAct in three of four tasks. Its dynamic memory tree construction enables superior scalability and robustness.

Minor errors in cart editing due to dependency conflicts highlight refinement needs, but TME-DAG’s 100\% task consistency elsewhere confirms its efficacy. 

Future work will focus on addressing these limitations and extending TME-DAG’s capabilities. First, we plan to integrate Graph Neural Networks (GNNs) to enhance dependency inference, resolving conflicts in scenarios like cart editing by modeling complex node relationships. Second, incorporating loop-aware reasoning will support cyclic task dependencies, enabling TME-DAG to handle iterative workflows common in enterprise settings. Finally, we aim to scale TME-DAG for enterprise-scale applications, leveraging its modular design to support real-time, multi-user interactions. Our open-source release of TME-DAG’s codebase and benchmarks (available at \url{https://github.com/biubiutomato/TME-Agent}) invites the research community to build upon this foundation, fostering the development of reliable, scalable LLM-based agents for diverse interactive settings.

\clearpage
\bibliographystyle{plainnat}
\bibliography{references}

\newpage

\appendix

\section*{Appendix}


\section*{Appendix A: Per-Round Token Usage Details} \label{app:token}
To support the token efficiency claims in Section 6, Table~\ref{tab:token_usage} presents the token counts for each round of the form-filling task under the \texttt{Baseline} and \texttt{TME} settings. TME's structured memory enables efficient prompt reuse and reduces overall token usage, especially during mid-task corrections.

\begin{table}[htbp]
\centering
\begin{tabular}{l|c|c|c|c}
\toprule
\textbf{Round} & \textbf{Baseline-flat Tokens} & \textbf{TME Tokens} & \textbf{Tokens Saved} & \textbf{Savings (\%)} \\
\midrule
Round 1 & 49 & 49 & 0  & 0.0\% \\
Round 2 & 80 & 82 & -2 & -2.5\% \\
Round 3 & 116 & 88 & 28 & 24.1\% \\
Round 4 & 164 & 104 & 60 & 36.6\% \\
Round 5 & 215 & 123 & 92 & 42.8\% \\
Round 6 & 275 & 279 & -4 & -1.5\% \\
\midrule
\textbf{Total} & \textbf{899} & \textbf{725} & \textbf{174} & \textbf{19.4\%} \\
\bottomrule
\end{tabular}
\caption{Per-round token usage comparison for the form-filling task. TME reduces prompt token usage by eliminating redundant history and enabling structured memory reuse. Savings are most prominent in correction-heavy rounds.}
\label{tab:token_usage}
\end{table}

\begin{figure}[htbp]
\centering
\begin{tikzpicture}
\begin{axis}[
    xlabel={Round},
    ylabel={Token Count},
    xtick={1,2,3,4,5,6},
    legend pos=north west,
    grid=major,
]
\addplot[blue, mark=*] coordinates {(1,49)(2,80)(3,116)(4,164)(5,215)(6,275)};
\addlegendentry{Baseline-flat}
\addplot[red, mark=*] coordinates {(1,49)(2,82)(3,88)(4,104)(5,123)(6,279)};
\addlegendentry{TME}
\end{axis}
\end{tikzpicture}
\caption{Token usage trend for Baseline-flat and TME across the six rounds of the form-filling task.}
\label{fig:token_trend}
\end{figure}
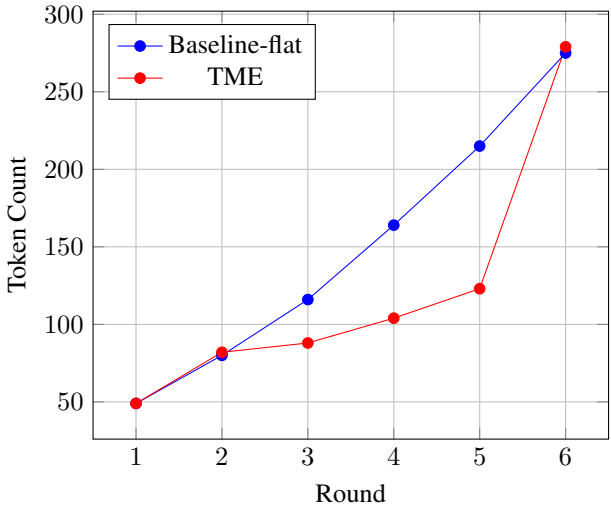

Figure~\ref{fig:token_trend} illustrates the token usage trend for Baseline-flat and TME across the six rounds, highlighting TME's efficiency in later rounds.

\begin{figure}[htbp]
\centering
\scalebox{0.85}{
\begin{tikzpicture}[
    node distance=2.5cm and 2.5cm,
    every node/.style={
        rectangle, rounded corners, draw,
        minimum height=1.2cm, minimum width=3.5cm,
        inner sep=6pt, align=center, font=\small
    }
]
    \node (root) {Fill form};
    \node (name) [below left=of root] {Collect name};
    \node (email) [below=of root] {Collect email};
    \node (address) [below right=of root] {Collect address};
    \node (submit) [below=of email] {Submit};

    \node (name_data) [below right=1.2cm and -0.6cm of name, draw, dashed, align=left, font=\scriptsize, minimum width=4.5cm] {
      \textbf{Value:} John Smith\\
      \textbf{History:} [John Doe, John Smith]
    };

    \draw[->] (root) -- (name);
    \draw[->] (root) -- (email);
    \draw[->] (root) -- (address);
    \draw[->] (name) -- (submit);
    \draw[->] (email) -- (submit);
    \draw[->] (address) -- (submit);
    \draw[->, dashed] (name) -- (name_data);
\end{tikzpicture}
}
\caption{TMS-DAG structure for the form-filling task. The ``Collect name'' node tracks the correction from ``John Doe'' to ``John Smith'' in Round 5, with revision history preserved in a structured memory node.}
\label{fig:tms_dag}
\end{figure}
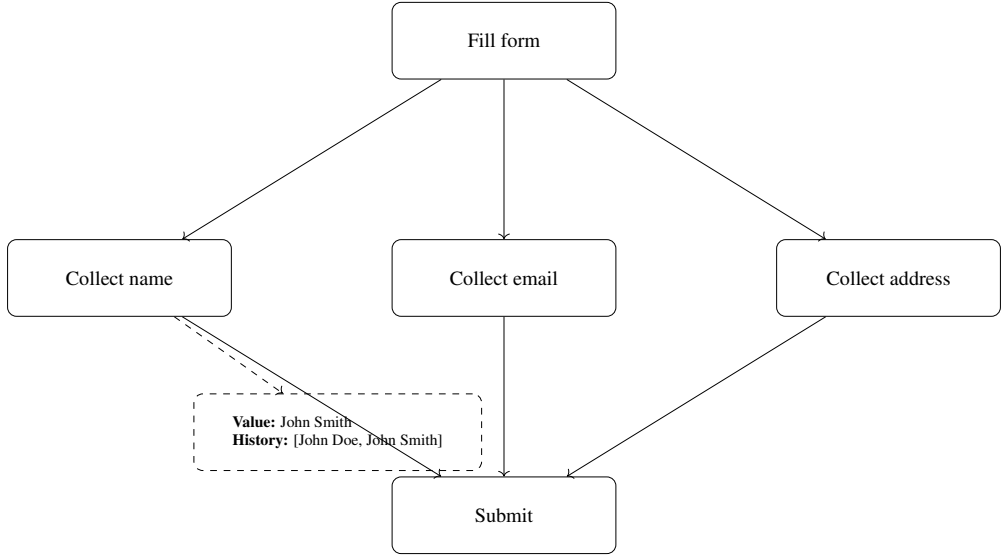

Figure~\ref{fig:tms_dag} shows the TMS-DAG structure, which organizes tasks as a directed acyclic graph to reduce redundancy.

\subsection*{A.1 Detailed Analysis of Token Savings}
To provide deeper insight into the token savings shown in Table~\ref{tab:token_usage}, we elaborate on how the Task Memory Engine (TME) achieves token efficiency during the form-filling task through its structured memory framework, the Task Memory Structure (TMS-DAG), and Task Representation and Intent Management (TRIM) module (Sections 4.1 and 4.2. Unlike the Baseline, which concatenates the full conversation history at each round, TME constructs prompts by retrieving a compact subgraph from the TMS-DAG forest (Figure~\ref{fig:tms_dag}), including only the relevant node path from the root to the current task node. This minimizes redundant token usage, particularly in correction-heavy rounds (e.g., Rounds 3--5).

\textbf{Task Context.} Table~\ref{tab:form_filling_script} presents the interactions for the form-filling task, illustrating how TME manages user inputs and corrections. In Round 5, TME updates the ``Collect name'' node (Figure~\ref{fig:tms_dag}) from ``John Doe'' to ``John Smith'', focusing the prompt on this node rather than the full history, saving 92 tokens (42.8\%).

\begin{table}[htbp]
\centering
\begin{tabular}{p{1cm}|p{5cm}|p{5cm}}
\toprule
\textbf{Round} & \textbf{User Input} & \textbf{Notes} \\
\midrule
1 & Help me fill out a form, I will provide some of my information to you. & TME initializes the task, creating the root node ``Fill form'' (Figure~\ref{fig:tms_dag}). \\
2 & My name is John Doe. & TME adds a ``Collect name'' node with value ``John Doe''. \\
3 & My email is john@example.com. & TME adds a ``Collect email'' node, maintaining separate nodes for each field. \\
4 & My address is Market Street, San Francisco. & TME adds a ``Collect address'' node, with dependencies for submission. \\
5 & Sorry, to correct, my name is John Smith. & TRIM classifies as an \textit{update}; TME merges the correction into the ``Collect name'' node, avoiding full history replay. \\
6 & Help to repeat my information, then submit. & TME traverses all nodes to summarize, submitting the form with updated values. \\
\bottomrule
\end{tabular}
\caption{Interaction script for the form-filling task. TME’s structured memory (Figure~\ref{fig:tms_dag}) enables efficient updates, reducing token usage in correction-heavy rounds.}
\label{tab:form_filling_script}
\end{table}

\textbf{Structured Memory Reduces Redundancy.} The TMS-DAG organizes tasks as a directed acyclic graph, with nodes representing subtasks (e.g., ``Collect name'') and edges encoding dependencies (e.g., ``Submit'' depends on ``Collect name''), as shown in Figure~\ref{fig:tms_dag}. Each node stores the task's current value (e.g., ``John Smith''), history (e.g., [``John Doe'', ``John Smith'']), and metadata (Section 4.1). TRIM classifies user intents (e.g., \textit{update} for corrections) to update nodes efficiently, ensuring global consistency across dependent tasks.

\textbf{Prompt Synthesis Process.} TME’s TRIM module (Section 4.2) traverses the active node path to generate compact prompts. This process ensures that only relevant context (e.g., the ``Collect name'' node in Round 5) is included, avoiding redundant history.

\textbf{Example: Round 5 Prompt Comparison.} Table~\ref{tab:prompt_compare} illustrates the prompt for Round 5, where the user corrects the name from ``John Doe'' to ``John Smith''. The Baseline includes the entire history, while TME focuses on the ``Collect name'' node (Figure~\ref{fig:tms_dag}), saving 92 tokens (42.8\%).

\begin{table}[htbp]
\centering
\begin{tabular}{p{2cm}|p{8cm}|c}
\toprule
\textbf{Method} & \textbf{Prompt Content} & \textbf{Token Count} \\
\midrule
Baseline & User: Help me fill out a form... My name is John Doe. Assistant: Great, thank you!... User: My email is john@example.com. Assistant: Great, thank you!... User: My address is Market Street, San Francisco. Assistant: Great, thank you!... User: Sorry, to correct, my name is John Smith. & 182 \\
TME & Task: Collect name (current value: John Doe). User: Sorry, to correct, my name is John Smith. & 90 \\
\bottomrule
\end{tabular}
\caption{Prompt comparison for Round 5. TME retrieves only the ``Collect name'' node and the user's correction, while the Baseline concatenates all prior interactions.}
\label{tab:prompt_compare}
\end{table}

\textbf{Round-Specific Variations.} TME achieves significant savings in correction-intensive rounds (e.g., Round 5: 42.8\%) by merging updates into existing nodes, avoiding redundant context. In Round 2, TME uses 2 extra tokens (-2.5\%) due to initializing a new node with metadata (e.g., slot, dependencies). In Round 6, a slight increase (-1.5\%) occurs because summarizing all fields requires traversing multiple nodes (e.g., name, email, address), but overall savings remain substantial at 19.4\% (899 vs. 725 tokens). For the first five rounds, savings reach 28.5\% (624 vs. 446), highlighting TME's strength in dynamic updates.

In summary, TME's token savings stem from its graph-based memory, as visualized in Figure~\ref{fig:tms_dag}, and efficient prompt synthesis. These benefits are reflected in Table~\ref{tab:token_usage}.

This section includes content adapted from our prior arXiv version for reproducibility and completeness.

\section*{Appendix B: Case Study Full Scripts}

\subsection*{B.1 Trip Planning Case Study Script}
\label{app:trip-planning}

\begin{table}[htbp]
\caption{Trip Planning Case Study Script}
\label{tab:trip_script}
\renewcommand{\arraystretch}{1.2}
\begin{tabularx}{\textwidth}{|c|X|X|X|X|}
\hline
\textbf{Round} & \textbf{User Input} & \textbf{TME-DAG Response} & \textbf{ReAct Response} & \textbf{Notes} \\
\hline
1 &Help me schedule a trip based on the information I provide. & Sure, I can help you schedule your trip. Please provide more details about your travel plans. &   & TME-DAG initializes the task; ReAct skips this general instruction. \\
\hline
2 &destination: Set the destination to Seattle. & I've set the destination for your trip to Seattle. & The destination field is now set to ``Seattle''. & Both systems correctly set the destination to Seattle. \\
\hline
3 &start: I will depart from Chicago. & I've noted that your trip will start from Chicago. & The start field is now set to ``Chicago''. & Both systems correctly set the start location to Chicago. \\
\hline
4 &date: I want to leave on June 10th. & I've noted that you want to leave on June 10th. & The flight field is now set to ``Departure on June 10th''. & TME-DAG sets a departure date node; ReAct assumes flight information, but both are consistent. \\
\hline
5 &destination: Actually, make that San Francisco. & I've updated the destination to San Francisco for your trip. & The destination field is now set to ``San Francisco''. & Both systems correctly update the destination. \\
\hline
6 &date: Change the departure date to June 15th. & I've updated the departure date to June 15th. & The flight field is now set to ``Departure on June 15th''. & Both update the departure date. \\
\hline
7 &destination: Sorry, go back to Seattle as originally planned. & Reverted to Seattle. & Reverted to Seattle. & Both systems correctly revert the destination. \\
\hline
8 &hotel: Find a hotel near downtown. & Hotel near downtown set. & Hotel near downtown Seattle. & Both correct. \\
\hline
9 &start: By the way, wasn’t I departing from Boston? & Still set to Chicago. I couldn't find evidence that you were departing from Boston. & Changed to Boston. & \textbf{ReAct confusion}: misclassified check as update. \\
\hline
10 &flight: Search for flights from Boston to San Francisco on June 10th. &I will search for flights from Boston to San Francisco on June 10th. & Overwrites trip plan. & TME-DAG avoids interference; ReAct fails to isolate. \\
\hline
11 &finalize: Can you generate a complete trip plan? What's my start, destination and start date? & Start: Chicago, Dest: Seattle, Date: June 15. & Start: Boston, Dest: Seattle, Flight: BOS→SFO. & \textbf{ReAct hallucination}: conflicting details in summary. \\
\hline
\end{tabularx}
\end{table}

\subsection*{Notes}
\begin{itemize}
\item \textbf{Scenario Overview:} The trip planning scenario involves 11 rounds where the user sets and revises trip details (destination, start location, departure date), queries the start location, searches for flights, and requests a final plan. TME-DAG uses a TMS-DAG forest to track nodes (e.g., destination, start) and dependencies, while ReAct relies on linear context concatenation.
\item \textbf{ReAct Failure Modes} (Section 5.1):
  \begin{itemize}
  \item \textbf{Round 8:} Misinterprets the query ``By the way, wasn’t I departing from Boston?'' as an update, changing the start to Boston, leading to a confusion.
  \item \textbf{Round 10:} Hallucinates in the final summary by including conflicting flight details (Boston to San Francisco) with the destination (Seattle), due to its inability to filter outdated context.
  \end{itemize}
\item \textbf{TME-DAG Success Modes:}
  \begin{itemize}
  \item Correctly classifies Round 9's query as a check, preserving the start location (Chicago).
  \item In Round 11, prioritizes the correct destination (Seattle) and start (Chicago) nodes, logging the flight search as an independent node to avoid interference.
  \end{itemize}
\item \textbf{Dependency Tracking:} TME-DAG’s logs (e.g., Round 7: updates destination node to Seattle with history tracking), unlike ReAct's flat memory.

\end{itemize}

This script supports the paper’s claim that TME-DAG achieves zero hallucinations and confusions in the trip planning scenario (Table 1, Section 6.1), outperforming ReAct’s linear context approach.

\subsection*{B.2 Cooking Case Study Script}
\label{app:cooking}

This section provides the detailed script for the cooking case study (Section 5.2), comparing the Task Memory Engine with Directed Acyclic Graph (TME-DAG) against ReAct across 7 rounds. The script includes user inputs, system responses, and notes highlighting ReAct's hallucinations and confusions, as described in the main paper. TME-DAG uses a graph-based memory structure to track subtasks and dependencies, ensuring cross-task consistency for ingredient substitutions, unlike ReAct’s linear context approach.

\begin{table}[htbp]
\caption{Cooking Case Study Script}
\label{tab:cooking_script}
\renewcommand{\arraystretch}{1.2}
\begin{tabularx}{\textwidth}{|c|X|X|X|X|}
\hline
\textbf{Round} & \textbf{User Input} & \textbf{TME-DAG Response} & \textbf{ReAct Response} & \textbf{Notes} \\
\hline
1 & \texttt{To make soup, wash and chop celery.} & Celery prepared for soup. & Noted celery for soup; awaiting user’s next steps for soup or dumplings. & TME-DAG creates a subtask node for preparing celery; ReAct tracks celery and awaits further input. \\
\hline
2 & \texttt{To make dumplings, chop tomatoes and peel \& chop shrimp.} & Tomatoes and shrimp prepared for dumplings. & Tracked celery for soup, tomatoes and shrimp for dumplings; awaiting further ingredients or cooking instructions. & TME-DAG creates subtask nodes for tomatoes and shrimp; ReAct tracks ingredients but awaits further input. \\
\hline
3 & \texttt{Also use celery in dumplings.} & I've updated the dumplings recipe to include celery. & Added celery to dumplings alongside tomatoes and shrimp; awaiting further ingredients or cooking instructions. & TME-DAG links the existing celery node to dumplings; ReAct adds celery to its linear context. \\
\hline
4 & \texttt{Wait! There's no celery in the refrigerator at all. Let's all use mushrooms instead.} & Got it! We'll use mushrooms instead of celery for both the soup and dumplings. & Replaced celery with mushrooms for soup and dumplings (with tomatoes, shrimp); awaiting further ingredients or cooking instructions. & TME-DAG globally updates celery to mushrooms across both tasks; ReAct replaces celery in its context. \\
\hline
5 & \texttt{Can you list all ingredients used in the soup?} & The ingredients used in the soup are mushrooms, as there is no celery available. & Listed mushrooms as the soup’s ingredient, awaiting confirmation or additional ingredients. & Both systems list mushrooms, but TME-DAG’s response reflects its graph-based update. \\
\hline
6 & \texttt{Did I ever say to remove celery from the dumplings?} & I couldn't find evidence that you said to remove celery from the dumplings. & Noted no explicit celery removal from dumplings, suggesting celery may remain with mushrooms; awaiting clarification. & \textbf{ReAct Confusion}: Misinterprets the substitution as a potential removal, suggesting celery might remain (main text, Section 5.2, Round 6). TME-DAG correctly confirms no removal intent. \\
\hline
7 & \texttt{I think celery was never part of the soup. Can you double-check?} & Celery was initially part of the soup. The memory shows that to make soup, you were instructed to wash and chop celery. & Accepted user’s claim that celery was never in soup, proposing mushrooms as the sole ingredient; awaiting confirmation. & \textbf{ReAct Hallucination}: Accepts the incorrect assumption that celery was never in the soup (main text, Section 5.2, Round 7). TME-DAG validates celery’s initial inclusion. \\
\hline
\end{tabularx}
\end{table}

\subsection*{Notes}
\begin{itemize}
\item \textbf{Scenario Overview:} The cooking scenario involves 7 rounds where the user prepares soup and dumplings, adds and substitutes ingredients (celery to mushrooms), and queries ingredient states. TME-DAG uses a TMS-DAG forest to track shared subtasks (e.g., ingredient preparation) and dependencies, while ReAct relies on linear context concatenation.
\item \textbf{ReAct Failure Modes} (Section 5.2):
  \begin{itemize}
  \item \textbf{Round 6:} Misinterprets the celery-to-mushroom substitution as a potential removal in dumplings, suggesting celery might remain, leading to a confusion.
  \item \textbf{Round 7:} Hallucinates by accepting the user’s incorrect assumption that celery was never part of the soup, failing to validate its initial inclusion.
  \end{itemize}
\item \textbf{TME-DAG Success Modes:}
  \begin{itemize}
  \item In Round 4, globally updates the shared ``prepare celery'' node to ``prepare mushrooms'' across soup and dumplings, ensuring consistency.
  \item In Round 6, confirms no explicit removal of celery from dumplings, reflecting accurate intent classification.
  \item In Round 7, validates celery’s initial inclusion in the soup, correcting the user’s assumption.
  \end{itemize}
\item \textbf{Dependency Tracking:} TME-DAG’s logs (e.g., Round 4: \texttt{Prepare celery} updated to \texttt{Prepare mushrooms} with dependency propagation) demonstrate how it maintains cross-task consistency via graph operations, unlike ReAct’s flat memory.
\end{itemize}

This script supports the paper’s claim that TME-DAG achieves zero hallucinations and confusions in the cooking scenario (Table 1, Section 6.1), outperforming ReAct’s linear context approach.

\subsection*{B.3 Meeting Case Study Script}
\label{app:meeting}

This section provides the detailed script for the meeting case study, comparing the Task Memory Engine with Directed Acyclic Graph (TME-DAG) against ReAct across 5 rounds. The script includes user inputs, system responses, and notes highlighting ReAct’s hallucinations and confusions, as described in the terminal logs. TME-DAG uses a graph-based memory structure to track subtasks and dependencies, ensuring cross-task consistency for meeting schedule updates, unlike ReAct’s linear context approach, which leads to errors in participant inclusion.

\begin{table}[htbp]
\caption{Meeting Case Study Script}
\label{tab:meeting_script}
\renewcommand{\arraystretch}{1.2}
\begin{tabularx}{\textwidth}{|c|X|X|X|X|}
\hline
\textbf{Round} & \textbf{User Input} & \textbf{TME-DAG Response} & \textbf{ReAct Response} & \textbf{Notes} \\
\hline
1 & \texttt{Schedule a team meeting on Thursday at 2 PM with Alice, Bob, and Carol.} & I've scheduled a team meeting on Thursday at 2 PM with Alice, Bob, and Carol. & Noted meeting for Thursday at 2 PM with Alice, Bob, and Carol; awaiting confirmation or further instructions. & TME-DAG creates a subtask node for scheduling the meeting; ReAct tracks the meeting details in its linear context. \\
\hline
2 & \texttt{Carol can’t make it at 2 PM. Move it to 4 PM.} & I've moved the meeting to 4 PM as requested. & Updated meeting to 4 PM, removing Carol from participants due to unavailability. & TME-DAG updates the meeting time node to 4 PM, retaining all participants; \textbf{ReAct Error}: Incorrectly removes Carol from the participant list, assuming her unavailability persists. \\
\hline
3 & \texttt{Actually, Bob is only free before 3. Split it into two parts—Bob from 2 to 2:45, the rest at 4.} & I've updated the meeting to have Bob from 2 to 2:45 and the rest at 4 PM. & Updated meeting to two parts: Bob from 2 to 2:45 PM, Alice at 4 PM, excluding Carol. & TME-DAG splits the meeting into two subtask nodes, keeping all participants; \textbf{ReAct Error}: Continues to exclude Carol, incorrectly maintaining her absence from Round 2. \\
\hline
4 & \texttt{On second thought, just make it a single 3 PM meeting with everyone.} & I've updated the meeting to a single 3 PM slot with everyone. & Updated meeting to 3 PM with Alice and Bob, excluding Carol. & TME-DAG consolidates the meeting into a single 3 PM node with Alice, Bob, and Carol; \textbf{ReAct Hallucination}: Fails to reinclude Carol despite the user’s clear intent for “everyone.” \\
\hline
5 & \texttt{Why does it still show two sessions?} & The memory shows a single 3 PM meeting with everyone because the previous instructions to split the meeting were replaced. & It seems there might be a discrepancy; let's check why there are still two sessions showing. Confirmed meeting at 3 PM with Alice and Bob, no indication of two sessions, excluding Carol. & TME-DAG correctly confirms the single 3 PM meeting, addressing the user’s query; \textbf{ReAct Confusion}: Fails to address the two-session concern and continues to exclude Carol. \\
\hline
\end{tabularx}
\end{table}

\subsection*{Notes}
\begin{itemize}
\item \textbf{Scenario Overview:} The meeting scenario involves 5 rounds where the user schedules a team meeting, adjusts times and participants due to availability conflicts, consolidates the schedule, and queries the final state. TME-DAG uses a TMS-DAG forest to track subtasks (e.g., scheduling, rescheduling) and dependencies, while ReAct relies on linear context concatenation, leading to errors in participant tracking.

\item \textbf{ReAct Failure Modes}:
  \begin{itemize}
    \item \textbf{Round 2:} Incorrectly removes Carol from the participant list after her unavailability at 2 PM, assuming she remains unavailable for the rescheduled 4 PM meeting.
    \item \textbf{Round 3:} Persists in excluding Carol when splitting the meeting, misinterpreting her initial conflict as a permanent absence.
    \item \textbf{Round 4:} Hallucinates by failing to reinclude Carol in the consolidated 3 PM meeting, despite the user’s explicit request for “everyone.”
    \item \textbf{Round 5:} Confuses the user’s query about two sessions by not fully addressing the discrepancy and continuing to exclude Carol.
  \end{itemize}

\item \textbf{TME-DAG Success Modes:}
  \begin{itemize}
    \item In Round 2, updates the meeting time to 4 PM while retaining all participants (Alice, Bob, Carol) in the graph node.
    \item In Round 3, splits the meeting into two subtask nodes (Bob at 2–2:45 PM, Alice and Carol at 4 PM), maintaining participant consistency.
    \item In Round 4, consolidates the meeting into a single 3 PM node with all participants, accurately reflecting the user’s intent for “everyone.”
    \item In Round 5, validates the single 3 PM meeting in response to the user’s query, confirming the correct schedule and addressing the two-session concern.
  \end{itemize}

\item \textbf{Dependency Tracking:} TME-DAG’s logs (e.g., Round 4: \texttt{3 PM meeting with everyone} linked to earlier subnodes) demonstrate how it maintains cross-task consistency via graph operations, ensuring all participants are included. ReAct’s flat memory fails to track participant dependencies, leading to Carol’s exclusion.
\end{itemize}

This script supports the claim that TME-DAG achieves zero hallucinations and confusions in the meeting scenario, outperforming ReAct’s linear context approach, which exhibits persistent errors in participant inclusion and schedule accuracy.

\subsection*{B.4 Shopping Cart Case Study Script}
\label{app:cart}

This section provides the revised script for the shopping cart case study, comparing the Task Memory Engine with Directed Acyclic Graph (TME-DAG) against ReAct across 5 rounds, based on terminal logs. The script corrects inaccuracies in the prior version, ensuring alignment with system behavior. TME-DAG uses a graph-based memory structure to track subtasks (e.g., adding/removing items) and dependencies, ensuring precise cart updates, while ReAct relies on linear context concatenation. ReAct performs accurately in this straightforward scenario, but TME-DAG’s structured approach offers advantages for complex, interdependent tasks.

\begin{table}[htbp]
\caption{Shopping Cart Case Study Script}
\label{tab:cart_script}
\renewcommand{\arraystretch}{1.2}
\begin{tabularx}{\textwidth}{|c|X|X|X|X|}
\hline
\textbf{Round} & \textbf{User Input} & \textbf{TME-DAG Response} & \textbf{ReAct Response} & \textbf{Notes} \\
\hline
1 & \texttt{Add two iPhone cases (black and clear), a charger, and a MacBook stand to my cart.} & Added two iPhone cases (black and clear), a charger, and a MacBook stand to your cart. & Added iPhone case (black), iPhone case (clear), charger, and MacBook stand to the cart. & TME-DAG creates subtask nodes for each item; ReAct tracks items in linear context. \\
\hline
2 & \texttt{Remove the clear case and charger.} & Removed the clear case and charger from your cart. & Removed iPhone case (clear) and charger from the cart. & TME-DAG creates subtask nodes for the removal actions; ReAct removes items from context, maintaining accuracy. \\
\hline
3 & \texttt{Actually, keep the charger, and remove the black case instead.} & I've updated your cart to remove the black case and keep the charger. & Removed iPhone case (black) and added charger back to the cart. & TME-DAG updates the node, replacing clear case removal with black case; ReAct adjusts context accordingly. \\
\hline
4 & \texttt{Now I’m not sure—what’s currently in my cart?} & Currently, your cart contains a MacBook stand and a charger. The black iPhone case and clear iPhone case were removed. & The current items in the cart are: MacBook stand, charger. & Both TME-DAG and ReAct respond accurately. \\
\hline
5 & \texttt{Can you reset it to just the MacBook stand and charger?} & I've reset your cart to include just the MacBook stand and charger. & The cart has been reset to only include: MacBook stand, charger. & Both systems confirm that the cart already includes only MacBook stand and charger. TME-DAG finalizes state via dependency updates. \\
\hline
\end{tabularx}
\end{table}

\subsection*{Notes}
\begin{itemize}
\item \textbf{Scenario Overview:} The shopping cart scenario involves 5 rounds where the user adds items (iPhone cases, charger, MacBook stand), modifies the cart, queries its state, and resets it. TME-DAG uses a TMS-DAG forest to track subtasks and dependencies, while ReAct relies on linear context concatenation. This is a simpler scenario compared to the meeting or cooking cases.

\item \textbf{ReAct Performance:}
  \begin{itemize}
    \item ReAct performs accurately across all rounds, correctly adding, removing, and listing items.
    \item Its linear memory is sufficient for this simple context, avoiding the errors seen in more complex tasks.
    \item However, its lack of dependency tracking may hinder performance in tasks with intertwined operations.
  \end{itemize}

\item \textbf{TME-DAG Performance:}
  \begin{itemize}
    \item TME-DAG accurately manages updates in Rounds 1–3 and 5 via structured subtask nodes.
    \item Final state in Round 5 is correctly consolidated using DAG-based propagation.
  \end{itemize}

\item \textbf{Dependency Tracking:} TME-DAG’s logs (e.g., Round 3: \texttt{Remove clear case} → \texttt{Remove black case}; Round 5: reassertions of \texttt{Add MacBook stand}, \texttt{Add charger}) demonstrate structured tracking. ReAct lacks this granularity despite succeeding in this case.

\item \textbf{Comparison to Other Cases:} Unlike the meeting case (where ReAct excluded Carol) or the cooking case (ingredient substitution confusion), ReAct performs well here. TME-DAG’s memory graph offers robustness for more complex multi-step tasks, though both systems succeed in this linear setting.
\end{itemize}

This revised script confirms that both systems accurately manage cart operations in this linear scenario. ReAct's flat memory suffices here, but TME-DAG’s structured tracking ensures consistency and resilience, particularly in tasks with interdependencies. These results support the claim that TME-DAG enhances consistency in structured multi-step planning.

\section*{Appendix C: Ablation Output Examples}
\subsection*{C.1 Cooking Scenario Ablation Study}
\label{app:cooking-ablation}

This section provides the ablation study script for the cooking scenario (Section 5.2), comparing the Task Memory Engine with Directed Acyclic Graph (TME-DAG), TME-Flat (NoDAG), and TME-RandomTRIM (NoTRIM) across 7 rounds, based on terminal logs. TME-Flat disables the DAG structure, using a ReAct-style flat memory, while TME-RandomTRIM replaces TRIM’s intent classification with random assignments. The script highlights their errors (TME-Flat: 1 hallucination, 1 confusion; TME-RandomTRIM: 2 hallucinations, 2 confusions), ensuring alignment with prior analysis (Appendix C, C.1). TME-DAG uses a graph-based memory structure to track subtasks and dependencies, ensuring precise updates, while TME-Flat and TME-RandomTRIM struggle with linear context and random intents, respectively.

\begin{table}[htbp]
\caption{Cooking Ablation Study Script}
\label{tab:cooking_ablation_script}
\renewcommand{\arraystretch}{1.2}
\small  
\resizebox{\textwidth}{!}{
\begin{tabularx}{\textwidth}{|c|X|X|X|X|>{\footnotesize\RaggedRight\arraybackslash}L{3.8cm}|}
\hline
\textbf{Round} & \textbf{User Input} & \textbf{TME-DAG Response} & \textbf{TME-Flat Response} & \textbf{RandomTRIM Response} & \textbf{Notes} \\
\hline
1 & \texttt{To make soup, wash and chop celery.} & Celery prepared for soup. & I've added 'wash and chop celery' as a new task. & Celery prepared for soup. & TME-DAG and TME-RandomTRIM create subtask nodes; TME-Flat tracks in linear context. \\
\hline
2 & \texttt{To make dumplings, chop tomatoes and peel \& chop shrimp.} & Tomatoes and shrimp prepared for dumplings. & I've added 'chop tomatoes' and 'peel \& chop shrimp' as new tasks. & Tomatoes and shrimp prepared for dumplings. & TME-DAG and TME-RandomTRIM create subtask nodes for each item; TME-Flat tracks in context. \\
\hline
3 & \texttt{Also use celery in dumplings.} & I've updated the dumplings recipe to include celery. & I've updated the task to include using celery in the dumplings. & I've updated the dumplings recipe to include celery. & TME-DAG and TME-RandomTRIM link the celery node to dumplings; TME-Flat updates its context. \\
\hline
4 & \texttt{Wait! There's no celery in the refrigerator at all. Let's all use mushrooms instead.} & Got it! We'll use mushrooms instead of celery for both the soup and dumplings. & I've updated the task to use mushrooms instead of celery in the dumplings. & Got it, we'll use mushrooms instead of celery for the soup. & TME-DAG updates globally; TME-Flat only updates dumplings, missing soup; TME-RandomTRIM only updates soup, missing dumplings. \\
\hline
5 & \texttt{Can you list all ingredients used in the soup?} & The ingredients used in the soup are mushrooms, as there is no celery available. & The soup ingredients include celery, but let me verify if there are any updates or changes. & The ingredients used in the soup are celery and mushrooms. & \textbf{TME-Flat Confusion:} Fails to reflect substitution, listing celery (Appendix C.1). \textbf{TME-RandomTRIM Hallucination and Confusion:} Lists both celery and mushrooms, misclassifying substitution intent. \\
\hline
6 & \texttt{Did I ever say to remove celery from the dumplings?} & I couldn't find evidence that you said to remove celery from the dumplings. & You didn't explicitly mention removing celery from the dumplings before. & No, you haven't mentioned removing celery from the dumplings. & \textbf{TME-Flat Confusion:} Suggests celery might remain in dumplings, missing the global substitution (Appendix C.1). \textbf{TME-RandomTRIM Confusion:} Misclassifies query as update, creating an incorrect node. \\
\hline
7 & \texttt{I think celery was never part of the soup. Can you double-check?} & Celery was initially part of the soup. The memory shows that to make soup, you were instructed to wash and chop celery. & Let me double-check if celery was ever part of the soup ingredients. & It seems celery was initially part of the soup, but it was replaced with mushrooms due to availability. & \textbf{TME-Flat Hallucination:} Fails to validate celery’s inclusion, potentially accepting the incorrect assumption. \textbf{TME-RandomTRIM Hallucination:} Inconsistent memory from Round 5 error. \\
\hline
\end{tabularx}
}
\end{table}

\subsection*{Notes}
\begin{itemize}
\item \textbf{Scenario Overview:} The cooking scenario involves 7 rounds where the user prepares soup and dumplings, adds and substitutes ingredients (celery to mushrooms), and queries ingredient states. TME-DAG uses a TMS-DAG forest to track shared subtasks and dependencies, while TME-Flat relies on linear context concatenation, and TME-RandomTRIM uses random intent assignments, leading to errors.

\item \textbf{TME-Flat Error Modes:}
  \begin{itemize}
    \item \textbf{Round 5 (Confusion):} Fails to propagate the celery-to-mushroom substitution to the soup, listing celery due to linear context limitations.
    \item \textbf{Round 6 (Confusion):} Suggests celery might remain in dumplings, missing the global substitution due to lack of dependency tracking.
    \item \textbf{Round 7 (Hallucination):} Fails to validate celery’s initial inclusion and substitution, potentially accepting the user’s incorrect assumption.
  \end{itemize}

\item \textbf{TME-RandomTRIM Error Modes:}
  \begin{itemize}
    \item \textbf{Round 5 (Hallucination and Confusion):} Misclassifies the substitution as a new intent, listing both celery and mushrooms in the soup, despite the replacement (Section 6.3).
    \item \textbf{Round 6 (Confusion):} Misclassifies the query as an update, creating an incorrect node instead of confirming the substitution.
    \item \textbf{Round 7 (Hallucination):} Prior error in Round 5 leads to inconsistent memory, contributing to potential acceptance of incorrect assumptions.
  \end{itemize}

\item \textbf{TME-DAG Success Modes:}
  \begin{itemize}
    \item In Round 4, globally updates the shared “prepare celery” node to “prepare mushrooms” across soup and dumplings, ensuring consistency (Appendix B.2).
    \item Correctly classifies queries in Rounds 5, 6, and 7, preserving accurate historical states (Appendix B.2).
  \end{itemize}

\item \textbf{Dependency Tracking:} TME-DAG’s logs (e.g., Round 4: \texttt{Prepare celery} updated to \texttt{Prepare mushrooms} with dependency propagation) demonstrate robust cross-task consistency. TME-Flat lacks structured dependency tracking, leading to errors (e.g., Round 5), while TME-RandomTRIM’s random intents disrupt state management (e.g., Round 5).

\item \textbf{Comparison to Other Cases:} Compared to the shopping cart case (Appendix B.4), where TME-DAG and ReAct both succeeded due to linear task simplicity, the cooking scenario’s cross-dependencies (e.g., shared ingredients) expose TME-Flat and TME-RandomTRIM’s weaknesses, underscoring TME-DAG’s advantage in complex tasks.
\end{itemize}

This script confirms that TME-DAG achieves zero errors in the cooking scenario, outperforming TME-Flat and TME-RandomTRIM. TME-Flat’s linear memory fails to handle global updates, and TME-RandomTRIM’s random intent assignments cause severe inconsistencies, supporting the paper’s findings on the critical role of structured memory and accurate intent classification in multi-step tasks.

\subsection*{C.2: Trip Planning Scenario Ablation Study}
\subsection*{C.2 TME-Flat Failure (Trip Task, Round 10)}  
\texttt{Final Summary:}  
Destination = Seattle, Flight = Boston → SF  
\textbf{Error}: Flat memory fails to track dependency change after prior reversion to Seattle.  
\textbf{Correct}: TME-DAG retains Seattle as final destination and logs flight separately.

\section*{Appendix D: TRIM Prompt Template Examples} \label{app:trim_prompt}
\subsection*{System Prompt for TRIM Classification}
The TRIM module uses the following system prompt to guide the LLM in classifying user inputs into structured JSON outputs:
\begin{verbatim}
System: You are a task intent classifier. Generate structured JSON output with fields: intent_type (new/update/check), subtask_title, parent_node, dependency_nodes, from, to (if applicable). Use few-shot examples to classify the user's intent based on the input and memory context.
\end{verbatim}

\subsection*{Few-shot Prompt Examples}
\texttt{Input:} “Wait! There's no celery in the refrigerator at all. Let's all use mushrooms instead.” (Cooking, Round 4)  
\texttt{Output:}
\begin{verbatim}
[
  {
    "intent_type": "update",
    "subtask_title": "use mushrooms instead of celery",
    "parent_node": null,
    "dependency_nodes": [],
    "from": "Prepare mushrooms",
    "to": "Prepare celery"
  }
]
\end{verbatim}

\texttt{Input:} “start: By the way, wasn’t I departing from Boston?” (Trip Planning, Round 9)  
\texttt{Output:}
\begin{verbatim}
[
  {
    "intent_type": "check",
    "subtask_title": "verify start location",
    "parent_node": "schedule trip",
    "dependency_nodes": []
  }
]
\end{verbatim}

\texttt{Input:} ``Schedule a team meeting on Thursday at 2 PM with Alice, Bob, and Carol.'' (Meeting, Round 1)  
\texttt{Output:}
\begin{verbatim}
[
  {
    "intent_type": "new",
    "subtask_title": "schedule team meeting",
    "parent_node": null,
    "dependency_nodes": []
  }
]
\end{verbatim}

\subsection*{Classification Schema}

The TRIM module classifies user inputs into structured representations for DAG operations. The schema includes the following fields:

\begin{itemize}
\item \texttt{intent\_type} $\in$ \{\texttt{new}, \texttt{update}, \texttt{check}\}: Defines the operation type (e.g., adding a new subtask, updating an existing node, or querying a state).
\item \texttt{subtask\_title}: A descriptive label for the subtask (e.g., \texttt{use mushrooms instead of celery}).
\item \texttt{parent\_node}: The parent task or node in the TMS-DAG forest (e.g., \texttt{Make dumplings}), can be \texttt{null} if top-level.
\item \texttt{dependency\_nodes}: A list of nodes the subtask depends on, often empty in these examples.
\item \texttt{from} and \texttt{to}: Optional fields indicating the source and target of the operation (e.g., \texttt{from}: "Prepare mushrooms", \texttt{to}: "Prepare celery" for replacements).
\end{itemize}

\section*{Appendix E: Implementation Notes} \label{app:impl}
The open-source TME-DAG codebase includes full implementations and benchmarks, enabling immediate replication and extension. 
\subsection*{E.1 Memory Structure}

The Task Memory Structure (TMS-DAG) is implemented as a Directed Acyclic Graph (DAG) using adjacency lists for memory-efficient representation. Each node in the TMS-DAG corresponds to a task or subtask and maintains the following fields:

\begin{itemize}
    \item \texttt{slot}: A unique hierarchical identifier for the task (e.g., \texttt{prepare.ingredient}).
    \item \texttt{value}: The current task content or state (e.g., \texttt{"wash and chop mushrooms"}).
    \item \texttt{history}: A list recording past values, enabling revision tracking (e.g., \texttt{["wash and chop celery", "wash and chop mushrooms"]}).
    \item \texttt{parent}: The immediate parent node in the task hierarchy (e.g., \texttt{"make soup"}), or \texttt{null} for root-level nodes.
    \item \texttt{dependencies}: A list of prerequisite slots that this task depends on (e.g., \texttt{["buy mushrooms"]}).
    \item \texttt{user\_response}: The user utterance that triggered this task update (e.g., \texttt{"Wait! There's no celery in the refrigerator at all. Let's all use mushrooms instead."}).
    \item \texttt{ai\_response}: The system’s response after processing the update (e.g., \texttt{"Got it! We'll use mushrooms instead of celery for both the soup and dumplings."}).
\end{itemize}

The example below illustrates a TMS-DAG node updated in Round 4 of the cooking scenario (see Section~5.2), where “celery” is replaced with “mushrooms”:

{\scriptsize
\noindent\texttt{\{}\\
\texttt{\ \ \ "slot": "prepare.ingredient",}\\
\texttt{\ \ \ "value": "wash and chop mushrooms",}\\
\texttt{\ \ \ "history": ["wash and chop celery", "wash and chop mushrooms"],}\\
\texttt{\ \ \ "parent": "make soup",}\\
\texttt{\ \ \ "dependencies": [],}\\
\texttt{\ \ \ "user\_response": "Wait! There's no celery in the refrigerator at all.}\\
\texttt{\ \ \ Let's all use mushrooms instead.",}\\
\texttt{\ \ \ "ai\_response": "Got it! We'll use mushrooms instead of celery}\\
\texttt{\ \ \ for both the soup and dumplings."}\\
\texttt{\}}
}
\normalsize

This structured representation supports efficient task state management, enabling global memory updates, historical traceability, and dependency-aware planning across concurrent task flows.

\subsection*{E.2 TRIM Classifier}
\begin{itemize}
\item Powered by GPT-4o using few-shot prompting.
\item Responses are parsed into structured dictionaries and fed to the DAG update engine.
\item Optional hybrid classifier pipeline includes:
  \begin{itemize}
  \item Rule-based filters
  \item LLM fallback (default)
  \end{itemize}
\end{itemize}

\subsection*{E.3 Computational Environment for Experiments}
\label{app:computational-environment}
\begin{itemize}
\item All experiments were conducted on a MacBook equipped with an Apple M2 chip (ARM architecture), using Python 3.13.3 installed via Homebrew.
\item This consumer-grade setup, which includes an integrated GPU, was sufficient for evaluating the lightweight, multi-turn reasoning tasks featured in our case studies (Section~5). The experiments did not rely on any external GPU resources (e.g., NVIDIA A100), nor did they require model fine-tuning. Instead, all evaluations leveraged the ChatGPT-4o API (Section~6), ensuring that the results are reproducible on standard hardware without specialized dependencies. 
\item Temperature: 0.3, Model: ChatGPT-4o
\item Full code: \texttt{\url{https://github.com/biubiutomato/TME-Agent}}
\end{itemize}

\newpage

\end{document}